\documentclass[lettersize,journal]{IEEEtran}
\usepackage{amsmath,amsfonts,amsthm,amssymb}
\usepackage{array}
\usepackage[caption=false,font=normalsize,labelfont=sf,textfont=sf]{subfig}
\usepackage{textcomp}
\usepackage{stfloats}
\usepackage{url}
\usepackage{verbatim}
\usepackage{graphicx}
\usepackage{cite}
\usepackage{caption}
\usepackage{tabularx}
\usepackage{subcaption}
\usepackage{hyperref}
\usepackage{dsfont} 
\usepackage{stmaryrd} 
\usepackage{algorithm,algcompatible} 

\usepackage{xcolor}

\newcommand{\suma}[2]{\sum_{#1}^{#2}}

\begin{document}

\title{Vibration Damping in Underactuated Cable-suspended Artwork - Flying Belt Motion Control}




\author{
\IEEEauthorblockN{Martin Goubej\\}
\IEEEauthorblockA{NTIS Research Center\\
University of West Bohemia\\
Pilsen, Czechia\\
mgoubej@ntis.zcu.cz\\}
\and
\IEEEauthorblockN{Second Author\\}
\IEEEauthorblockA{Department of Test\\
Another University\\
City, Country\\
email@test.com}

\author{%
\IEEEauthorblockN{Martin Goubej\IEEEauthorrefmark{1},~\IEEEmembership{Member,~IEEE},
Lauria Clarke\IEEEauthorrefmark{2}, Martin Hraba\v{c}ka\IEEEauthorrefmark{1}, David Tolar\IEEEauthorrefmark{1} \\}
\IEEEauthorblockA{\IEEEauthorrefmark{1}NTIS Research Center, University of West Bohemia,
Pilsen, Czechia\\
Email: mgoubej@ntis.zcu.cz, dtolar@ntis.zcu.cz, hrabackm@kme.zcu.cz\\
}
\IEEEauthorblockA{\IEEEauthorrefmark{2}Antimodular Research Inc., Montreal, Canada \\
Email: lauria@antimodular.com}

}

\IEEEauthorblockA{\IEEEauthorrefmark{3}Department of Mechanics, Faculty of Applied Sciences, University of West Bohemia, Pilsen, Czechia\\
Email: hrabackm@kme.zcu.cz}

}



\markboth{IEEE Robotics \& Automation Magazine, 2026, AUTHOR ACCEPTED MANUSCRIPT}%
{Shell \MakeLowercase{\textit{et al.}}: A Sample Article Using IEEEtran.cls for IEEE Journals}


\maketitle

\begin{abstract}
This paper presents a comprehensive refurbishment of the interactive robotic art installation \textit{Standards and Double Standards} by Rafael Lozano-Hemmer. The installation features an array of belts suspended from the ceiling, each actuated by stepper motors and dynamically oriented by a vision-based tracking system that follows the movements of exhibition visitors. The original system was limited by oscillatory dynamics, resulting in torsional and pendulum-like vibrations that constrained rotational speed and reduced interactive responsiveness. To address these challenges, the refurbishment involved significant upgrades to both hardware and motion control algorithms. A detailed mathematical model of the flying belt system was developed to accurately capture its dynamic behavior, providing a foundation for advanced control design. An input shaping method, formulated as a convex optimization problem, was implemented to effectively suppress vibrations, enabling smoother and faster belt movements. Experimental results demonstrate substantial improvements in system performance and audience interaction. This work exemplifies the integration of robotics, control engineering, and interactive art, offering new solutions to technical challenges in real-time motion control and vibration damping for large-scale kinetic installations.
\end{abstract}


\section{Introduction}


\IEEEPARstart{T}he origins of robotic art can be traced from the automatons and mechanical toys of antiquity through to the mid 1950s, when consumer electronics and manufacturing robots became an exciting new medium for artists \cite{riskin}. Early work like Nicolas Schöffer's CYSP 1 (1956) -- a cybernetic sculpture that responded to viewer presence through sensors and analog electronics \cite{Schoffer63} -- established a template for contemporary robotic art, embedding electronic control systems and machines within artistic discourse. 


Today, contemporary artists continue to explore robotic technologies in their work. From the use of state-of-the-art manufacturing and aerial robots in choreographed performances, special-purpose machines designed from the ground up, to immersive, interactive robotic installations, robotic art has grown in variety and impact alongside the technologies it leverages \cite{kac1997digital, cuan2021, Zhang14}. Through these varied approaches, artists may focus on achieving specific aesthetic and conceptual goals, emphasizing emotional and physical interplay, or dissolving boundaries between observer and artwork -- transforming audiences from passive viewers to active participants \cite{Qin_2025}. In such contexts, technology becomes a tool to create moments of critical engagement for viewers.  

From an engineering perspective, robotic art installations face technical challenges that intersect mechanics, control systems, and human-robot interaction.
To sustain engaged viewer interactions in applications like collaborative drawing or audience-triggered motion, real-time control techniques that allow for dynamic responsiveness are essential to installation design \cite{gomez2021robot}. Building nuanced artistic expression with robotic components often requires advanced path-planning and motion control algorithms that account for machine and tool dynamics \cite{beltramello2020artistic}. Vision and motion-capture systems enable robots to perceive, interpret, and interact with their environment and audiences in sophisticated ways \cite{robinson2023robotic}. Robotic art has become a~legitimate field of research and practice within the engineering community, emphasizing the importance of interdisciplinary collaboration between artists, engineers, and designers \cite{st2019robotic}.

\begin{figure}[!t]
\centering
\includegraphics[width=0.45\textwidth]{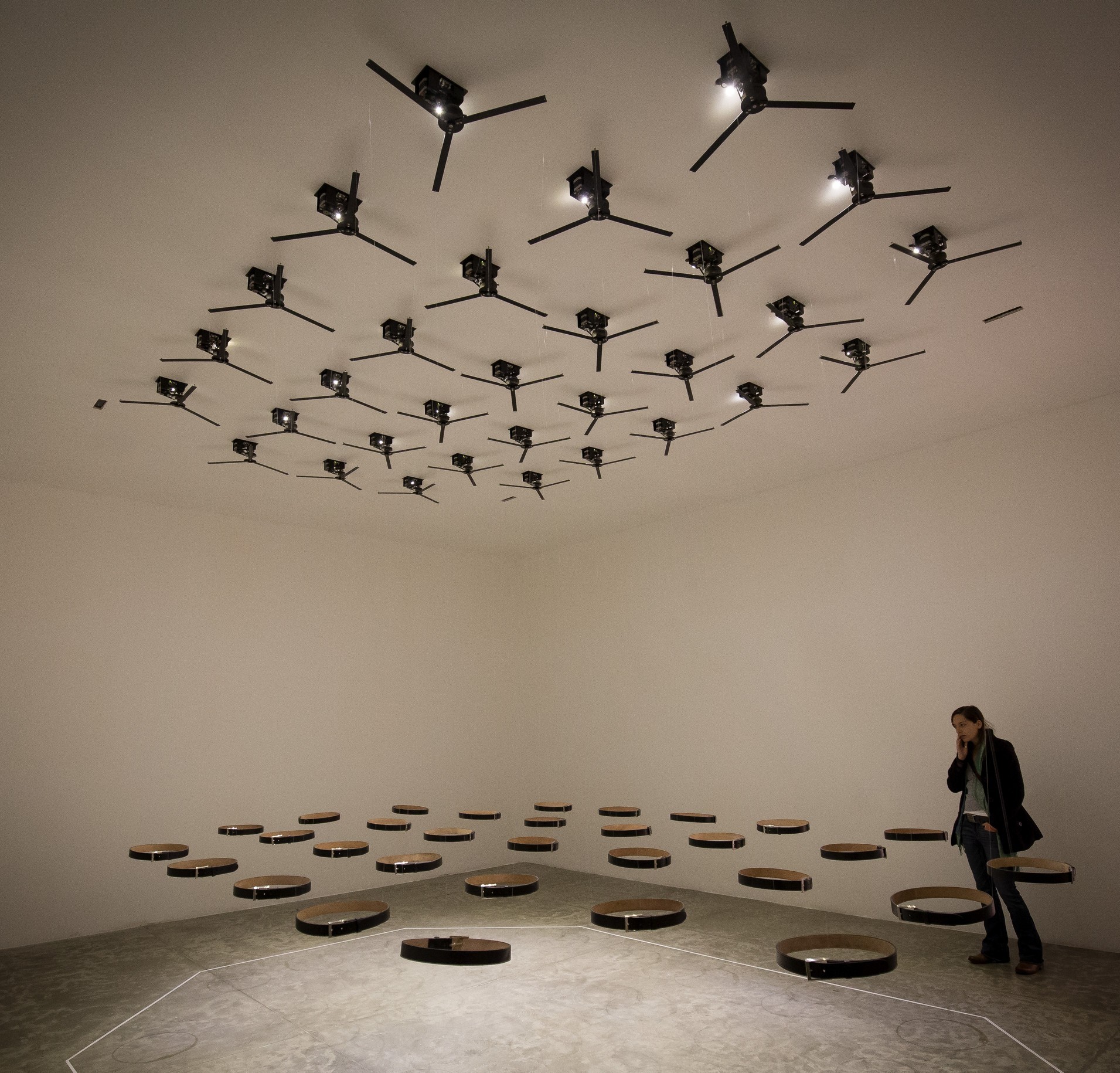}
\caption{Standards and Double Standards interactive installation}
\label{fig_standards}
\end{figure}

NOTE: This work has been accepted for publication in the IEEE Robotics and Automation Magazine under Creative Commons Attribution 4.0 International (CC-BY-4.0) license. Full version of the manuscript will be available via DOI \href{dx.doi.org/10.1109/MRA.2026.3687282}{10.1109/MRA.2026.3687282}. 

\newpage

This paper examines \textit{Standards and Double Standards}, an interactive installation created by Rafael Lozano-Hemmer, a~Mexican-Canadian media artist internationally recognized for his large-scale works that merge technology, architecture, and performance art \cite{standards}. 

The first version of the system, developed in 2004 in collaboration with Conroy Badger, comprised 10 to 100 waist-height belts, each suspended from the ceiling via thin filament and actuated by individual stepper motors (Fig. \ref{fig_standards}). The belts’ rotational movements were governed by a~computerized vision-based tracking system that continuously monitored and followed the positions of visitors within the exhibition space, creating an interaction between the observers and the observed objects. The technical components of the system were developed by Antimodular Research, a Montreal-based R\&D studio founded by Lozano-Hemmer as his main workplace for the design, engineering, and fabrication of large-scale interactive art installations.

The original system was constrained by limited rotational velocity and acceleration of the actuated belts, primarily due to oscillatory dynamics introduced by the flexibly suspended loads. This manifested as undesirable torsional and pendulum-like oscillations, which impaired precise motion control. To address this, the motion planning software implemented gradual acceleration and deceleration (ramp) profiles, partially suppressing oscillations but at the expense of reduced responsiveness to nearby participants. This compromise diminished the intended level of interactivity envisioned by the artist.

Two decades later, Antimodular Research undertook a comprehensive refurbishment of the installation, upgrading both the hardware instrumentation and motion control algorithms. In collaboration with the Automation, Control, and Robotic Systems group at the NTIS Research Center in Czechia \cite{across}, the project focused on increasing system speed and effectively mitigating unwanted oscillatory modes, thereby enhancing both performance and interactive experience.


The main contributions of this paper can be summarized as follows:
\begin{itemize}
    \item \textbf{C1} Overall architecture of the kinetic installation is presented, addressing both hardware and software components
    \item \textbf{C2} Mathematical model of the flying belt system is established, forming a baseline for numerical simulations and model-based design of motion control algorithm
    \item \textbf{C3} A solution for vibration damping is proposed, employing input shaping method embedded in the motion planning software of the robotic manipulators
\end{itemize}

Section II explains the background of the flying belts project from the artist's perspective. Section III provides an overview of the hardware and software building blocks of the refurbished system, giving a more technical insight and constituting the contribution C1. Section IV, related to C2, presents a detailed mathematical model of the flying belt system. Section V introduces the input shaping method, addressing C3.  Simulation and experimental results follow in Sections VI and VII.

\section{Project background}

\textit{Standards and Double Standards} is an early example of Lozano-Hemmer's approach to kinetic art as a means of exploring surveillance and interaction. The belt, a symbol of authority and institutional power, is recontextualized as an interactive element that both mirrors and questions systems of observation and control. This reflects Lozano-Hemmer's broader interest in surveillance technology as both a subject of political critique and a tool for creating responsive environments. The work functions on both individual and collective levels – single viewers trigger direct responses from the belts, while multiple participants generate complex patterns of movement. This scaling from individual to group interaction is characteristic of Lozano-Hemmer's installations where repeated elements combine, pushing viewers to consider the relationship between their own experience of an artwork and the collective behavior created as a result of many interactions.

\section{System overview}

The refurbished system can be broken into two components: the computer vision system, which tracks participants within the exhibition space, and the motion control system, which transforms commands from the vision system into rotational movement. Both aspects of the system were redesigned with flexibility, cost, and ease of conservation in mind. We were particularly focused on making the system flexible enough to be installed and reinstalled in different venues. This required us to create a simple yet effective interface for seamless reconfiguration of the motion planning algorithm for installation. 

\subsection{Hardware Components}

\begin{figure}[!t]
\centering
\includegraphics[width=0.485\textwidth]{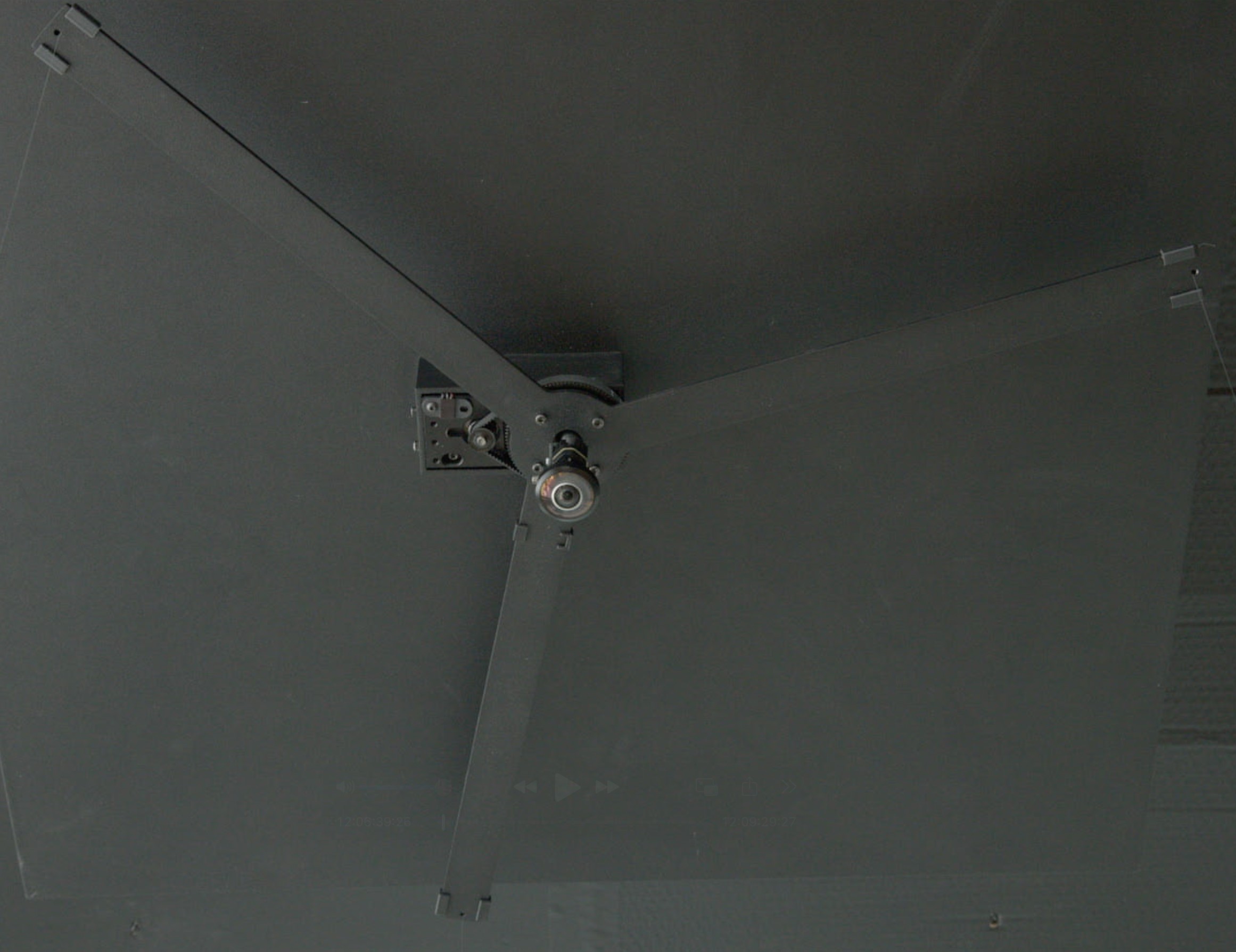}\\
\includegraphics[width=0.485\textwidth]{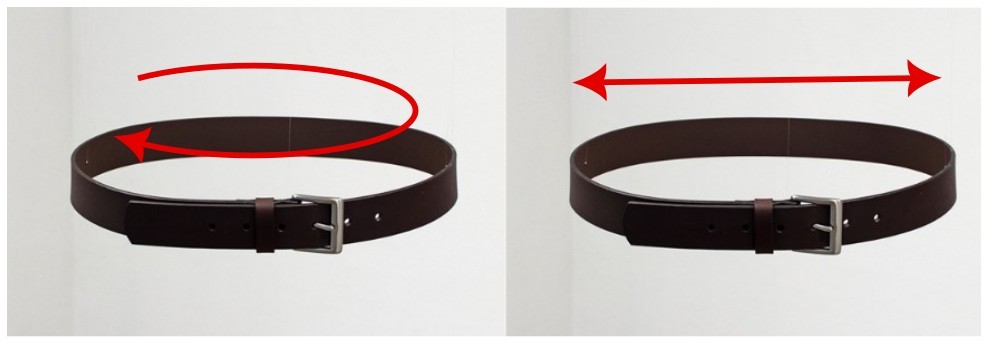}
\caption{Closeup view of mechanism showing rotary stage with camera and attachment cleats (top), detail of the suspended belt (bottom) with two fundamental modes of oscillation -- torsion (left) and nutation (right)}
\label{system_hw_real}
\end{figure}

\begin{figure}[!t]
\centering
\includegraphics[width=0.485\textwidth]{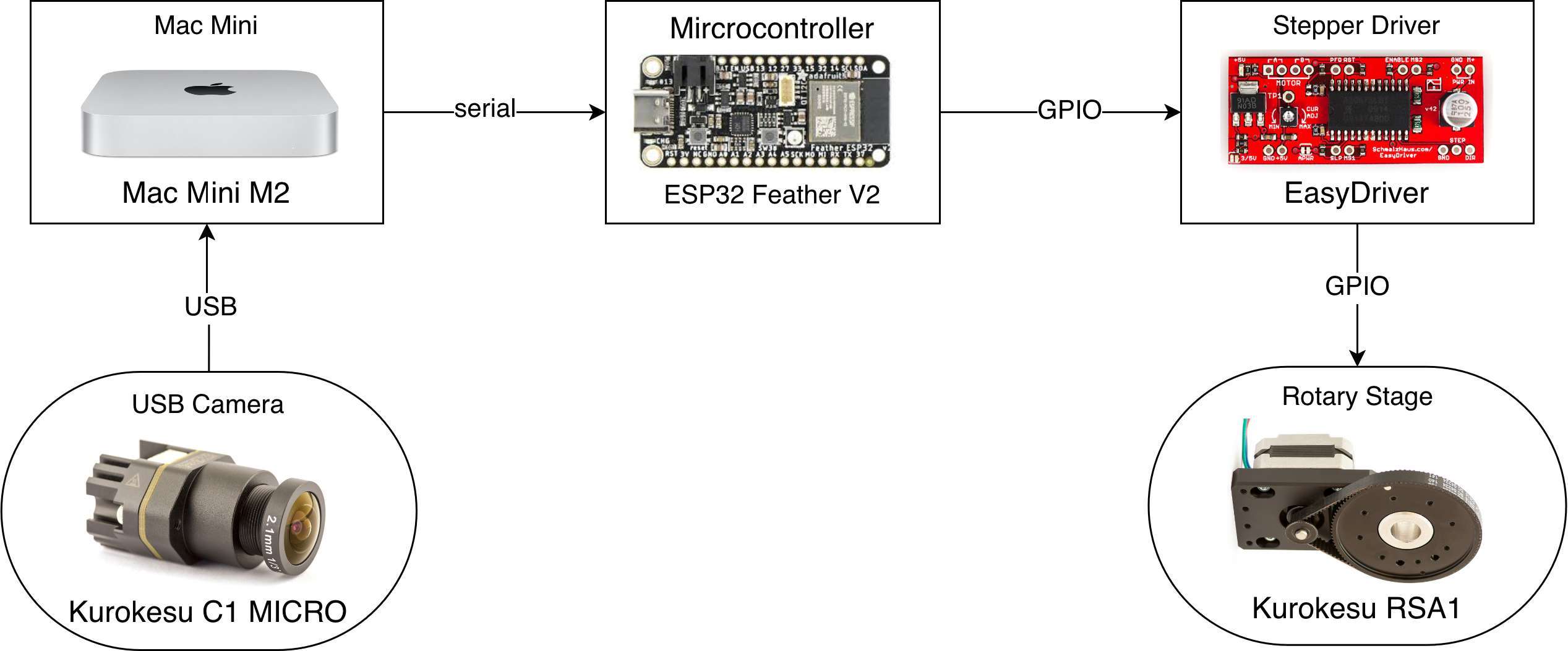}
\caption{Hardware diagram of the system}
\label{system_hw}
\end{figure}

The single belt version of the artwork consists of five main components (Fig. \ref{system_hw}). A generic USB webcam with a 180\textdegree  ~lens is used to capture a wide-angle, overhead view of the installation space. This camera feed is connected to a Mac Mini which runs a custom computer vision application tracking viewers as they move around the space. The Mac Mini sends trajectory update commands to an ESP32 microcontroller over a standard serial connection. The microcontroller applies multiple transformations to the computed trajectory before outputting GPIO signals to a stepper driver which controls the movement of a 1:5 geared rotary stage. Custom-built rotational arms are mounted to the rotary stage, allowing for easy suspension of the belt using a low-test monofilament line. 3D printed brackets were created to cleat the monofilament to the rotational arms, facilitating installation and adjustment of the belt's balance. 

Because the previously existing copy of the artwork belongs to a private collection, this iteration was redesigned to be scalable in a more flexible and cost-effective way. First, the motion control components used needed to be affordable and generic. We chose to use a simple GPIO-controlled stepper driver based on the A3967 chip manufactured by Allegro MicroSystems and assembled by SparkFun Electronics. It is often used in DIY and hobby electronics projects and provides a wide range of features useful for our application, such as micro-stepping. We chose to pair the motor driver with a~minimalistic belt-driven rotary stage actuated by a standard NEMA 17 bipolar stepper motor. This combination allows for up to 10,000 steps per revolution of control, or an angular resolution of 0.036 degrees. We also integrated an external magnetic hall sensor into the rotary stage which was triggered by a small magnet mounted on the rotary arm, allowing for precise homing of the manipulator.

\begin{table}
\centering
\begin{tabular}{|l|l|l|l|}
\hline
Component & Use & Manufacturer &  Price \\
\hline
\hline
C1 MICRO Camera & USB web cam & Kurokesu & \$130 \\
\hline
ESP32-PICO-MINI & motion control and & & \\ 
 & communication & Adafruit & \$20 \\
\hline
RSA1 Rotary Stage & rotary stage & Kurokesu & \$130 \\
\hline
EasyDriver & stepper control & SparkFun & \$18 \\
\hline
3D Printed Parts & mechanical assembly & Antimodular & \$5 \\ 
\hline
Total Cost & & & \$210 \\
\hline
\end{tabular}
\caption{Hardware components and cost (USD)}
\end{table}

Second, we wanted to ensure that the computational components chosen could support a wide range of application styles as we developed new software for the artwork. For the embedded motion control of the rotary stage we chose to use an ESP32-PICO-MINI based development board due to its multi-core processing, onboard flash memory, and networking capabilities. We chose to use a Mac Mini with an M2 processor to ensure high performance as we redesigned our computer vision application to leverage more up-to-date techniques. The use of the Mac Mini also grants us future flexibility with this application as we consider the integration of multiple camera views for a multi-instance installation of the artwork in a~larger space.  In a multi-instance installation, a single Mac Mini is capable of coordinating the movements of many ESP32 controlled rotary stages either via a wireless MQTT network, or a custom CAN topology currently under development by the authors. This topology also allows for ``over the air'' firmware updates to be propagated from the Mac Mini to each ESP32 controlled rotary stage in the installation. 

While more sophisticated combinations of rotary stage and motor controller would have afforded us more precise control over the rotation of the belt, we opted to maintain a low price point for scalability -- i.e. increasing the number of units (microcontroller and rotary stage) in the refurbished artwork -- and more generic components for ease of conservation.

\subsection{Software Components}

\begin{figure}[!t]
\centering
\includegraphics[width=0.485\textwidth]{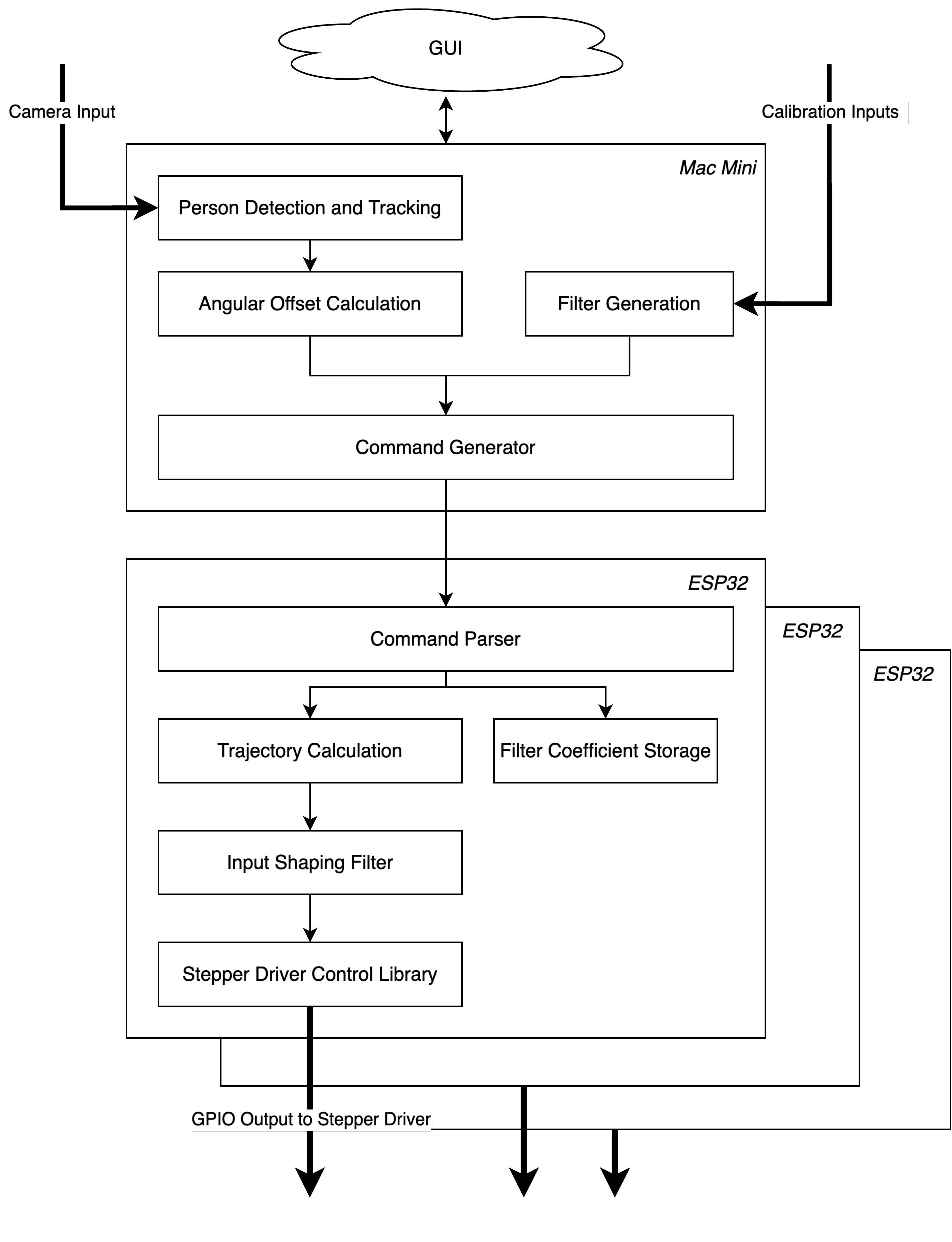}
\vspace{-1cm}
\caption{Software diagram of the system with multiple ESP32 motion control units.}
\label{system_sw}
\end{figure}

The software design of the system consists of three components: a computer vision-based overhead tracking system that calculates rotational offsets based on viewer position, motion control firmware that generates compensated motion patterns to achieve the target trajectory with minimal oscillation, and a~calibration tool for configuring the input shaping filter coefficients. The system was designed to be hardware-agnostic, with the tracking system decoupled from the motion control system, allowing multiple instances of the artwork to be integrated with a single instance of the vision system. We also focused on creating an intuitive configuration GUI so that non-technical installers can configure and calibrate the artwork on site. 

\begin{figure}[t]
\centering
\includegraphics[width=0.485\textwidth]{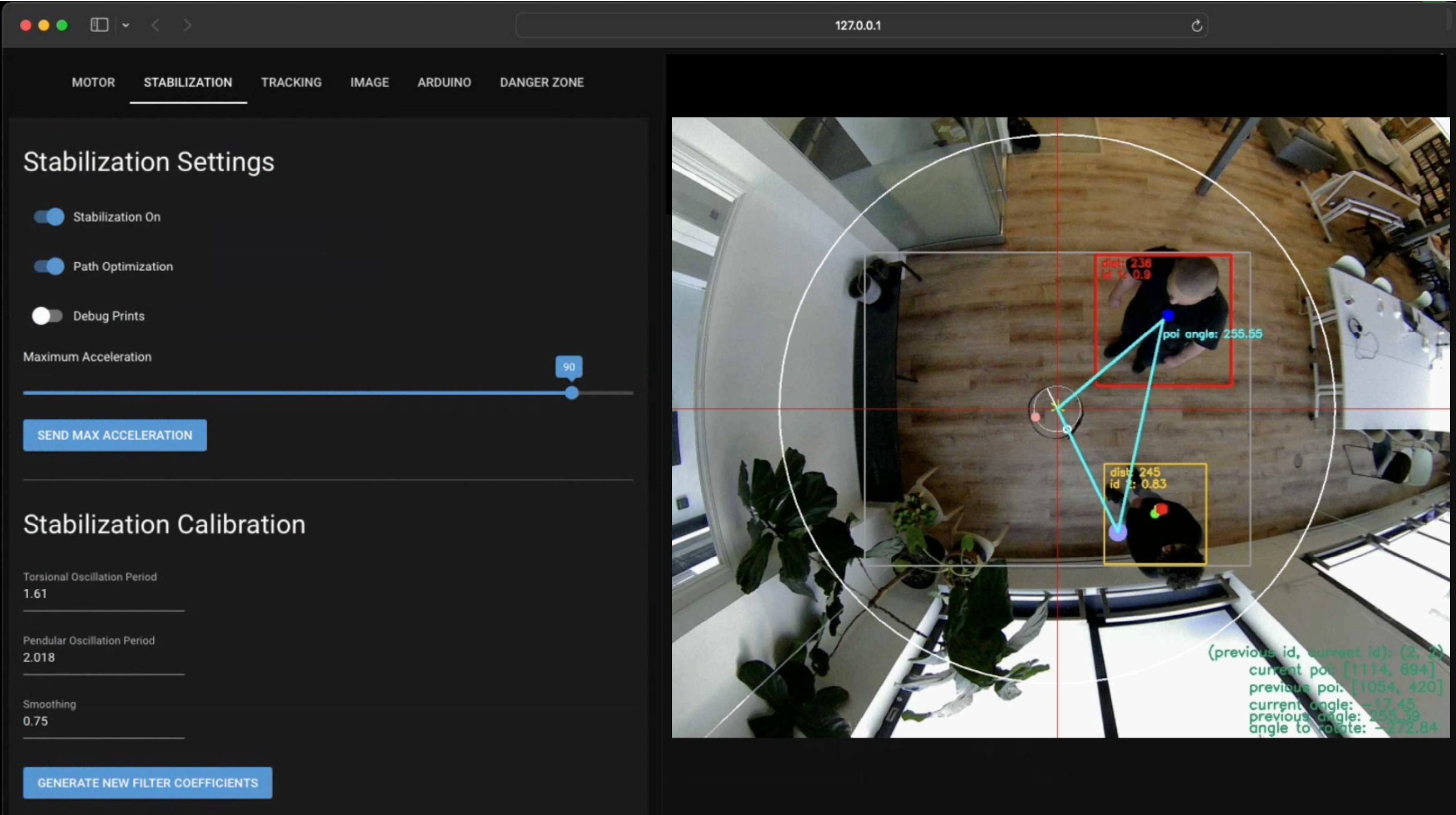}
\caption{Software interface showing vision system and stabilization settings}
\label{screenshot_sw}
\end{figure}

The computer vision system is written in Python with OpenCV and integrates a standard YOLO v8 object detection and tracking model using ByteSort with a GUI-based configuration system (Fig. \ref{screenshot_sw}), continuous angular offset calculations, and a~visual monitor window. The motion control system is written in C++ using the Arduino framework.

To configure and run the artwork, an installer first launches the software and begins the calibration process. After recording an average value for the pendular and rotational oscillation periods, these values are entered into the interface. An optimization algorithm described in Section V performs automated shaping filter synthesis resulting in an XML based coefficient file generated as an output. These coefficients are uploaded to the microcontroller via serial command, where they are stored in flash memory. Next, the installer configures the camera view and region of interest within the installation space. The field of view and region of interest can be adjusted based on the desired interaction range. After the belt is calibrated and the camera parameters are set, the rotary stage will begin its homing sequence. After locating the magnetic sensor, the piece returns to an idle location until a person appears in the field of view. The idle location and delay before returning to idle can be configured using the interface. The installer can then enable overhead tracking and normal operation begins.

When the computer vision system detects a person in the scene, the difference in their angular position from the belt's home location and distance from the belt's center is computed. This new angular offset is compared to the last angular position recorded, and if the change is sufficiently large, it is sent via serial to the microcontroller. The vision system will always choose to move toward the person whose bounding box has the shortest distance to the center of the belt if multiple people are detected in the scene. 

After the positional update command has been sent over serial, it is parsed by the microcontroller and the angular value is converted into an absolute offset in steps. This value is fed to a limiter function which clamps the value within the range of possible step values accounting for the shortest path from one rotational offset to another. The clamped value is then passed through an input shaping filter, whose parameters are designed by means of an optimization algorithm presented in Section V, to achieve a~modulated positional update value to effect a~vibration damped movement. The new position value is passed to a~generic stepper control library, AccelStepper, which actuates the driver to move to the desired location using a smoothed acceleration curve. This positional update process happens at a frequency of 100 Hz so that new input to the system results in real-time reaction speeds to engage viewers.

Both parts of the software system were designed with conservation and improvement of the artwork in mind; the vision system uses a free and generic object detection model operating on a standard camera feed to track people within the installation space, while the motion control system pairs our custom input shaping library with an open-source stepper control library. While these two key components are highly functional as-is, the software system is structured in such a~way that they can easily be replaced over time as use case and compatibility dictate.



\setlength{\unitlength}{1cm}
\begin{figure}[!t]
    \centering
    \subfloat[][\normalfont{Whole system -- 3D view} \label{fig_mat_model_3D}]
        {\begin{picture}(7,5)
        \put(-0.95,0){\includegraphics[width=0.48\textwidth]{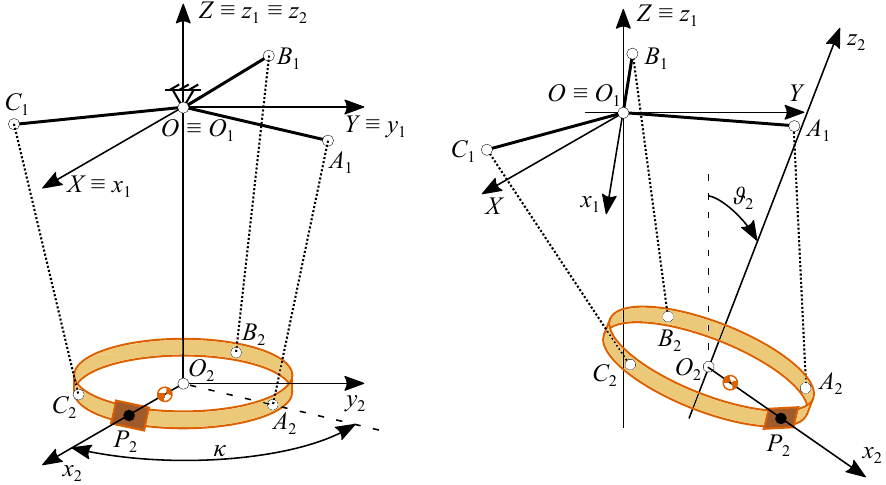}}
        \end{picture}} \quad
    \subfloat[][\normalfont{Body 2 -- 2D upper view} \label{fig_mat_model_2D}]
        {\begin{picture}(7,3)
        \put(-0.95,0){\includegraphics[width=0.48\textwidth]{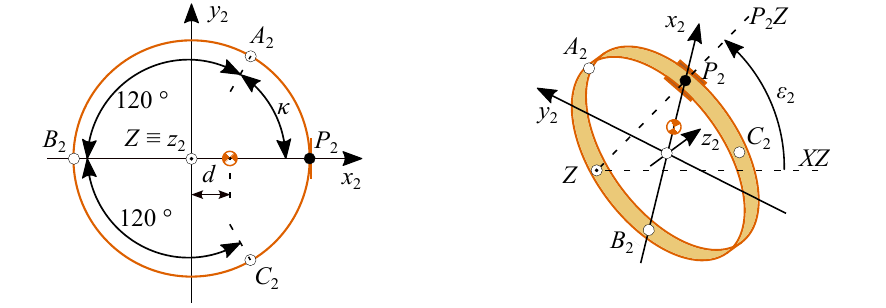}}
        \end{picture}}
    \caption{The modeled system in the initial configuration $t=0$ (left) and in a general configuration $t>0$ (right)}
    \label{fig_mat_model}
\end{figure}

\section{Dynamic Model of the Belt System}


\noindent We start with the following modeling assumptions:
\begin{itemize}
\item Cable stiffness/elasticity -- the cables are made from a~high-strength filament and are preloaded due to the gravity acting on the suspended belt. During normal system operation, there are no sudden movements of the actuator that would cause cable deformation or oscillation. Therefore, we assume a constant distance between the cable's connecting points as a fixed kinematic constraint.
\item Belt flexibility -- due to the relatively low rotational velocity and acceleration, there are no significant inertial forces that would cause belt deformation and we assume it to be a~rigid body.
\item The dynamics of the motor, occurring on a significantly faster time scale than the mechanics, is neglected, and the actuator is modeled as an ideal revolute joint.

\end{itemize}
Using these assumptions, the system can be modeled as a constrained multibody system (Fig. \ref{fig_mat_model}) comprising the Motorized Cable Suspension Unit (MCSU), denoted as body 1, and the belt with a metal buckle, denoted as body 2 and colored orange for clarity.  The MCSU consists of three straight anchor arms of the same length $r_1$, all lying in one plane and forming an angle of 120 deg with the neighbor anchor. The MSCU is connected in its center to a~stepper motor controlling the orientation of the MSCU. A cable connecting the MCSU and the belt is attached to each end of the anchor arm. The belt is circular with the radius $r_2$, and cables are evenly distributed along the circumference. There is a metal buckle that highly influences the dynamics of the system, so it is necessary to include it in the model. The buckle's position $P_2$ is determined by the offset angle $\kappa$ defined according to Figures \ref{fig_mat_model_3D} and \ref{fig_mat_model_2D}.


Each body has its local coordinate system defined by the referential point $O_i$, and three perpendicular axes $x_i, y_i, z_i$, where $i$ is the body index. Referential points are attached to geometrical centers of respective bodies, the axis $x_2$ passes through the point $P_2$ representing the buckle on the belt, and the axis $x_1$ is oriented similarly in the initial configuration ($t=0$). There is also a global coordinate system, which is fixed in space and represents the frame. Its referential point is denoted as $O$ and three perpendicular axes are denoted as $X, Y, Z$. In the initial configuration $t=0$, the global coordinate system has the same position and orientation as the local coordinate system of body 1 (see Figure \ref{fig_mat_model_3D} left).

Mass properties of the body are expressed by the 6-by-6 diagonal mass matrix $\mathbf{M}_i$ related to the center of mass of the body $i$, and the position of the center of mass in the local coordinate system. The center of mass of body 1 is identical to the referential point $O_1$, and the center of mass of body 2 lies on the axis $x_2$ in the distance $d$ from the referential point $O_2$.

Body 1 is attached to the frame by a revolute joint, enabling single rotation along the vertical axis $z_1$ and no translation. Therefore, this body has one degree of freedom. Additionally, there are three cables connecting both bodies together. Considering the initial assumption that cables are rigid (no elastic behavior) and they are loaded only in tension (no pressure), each cable eliminates one degree of freedom, and body 2 has three degrees of freedom as a consequence. In total, the presented system has four degrees of freedom.

\subsection{First principle model}

Dynamics of a multibody system can be described by the set of differential-algebraic equations (DAEs) of index one written in the simplified matrix form as
\begin{equation}
\begin{bmatrix}
\mathbf{M} & {\partial_\mathbf{q}\mathbf{c}}^\mathrm{T}\\
\partial_\mathbf{q}\mathbf{c} & \mathbf{0}
\end{bmatrix}
\begin{bmatrix}
\mathbf{\ddot{q}}\\
\boldsymbol{\lambda}
\end{bmatrix}
= \begin{bmatrix}
\mathbf{f}^E+\mathbf{f}^V\\
\mathbf{f}^D
\end{bmatrix},
\label{eq:mod_1}
\end{equation}
where $\mathbf{M}$ is the mass matrix, $\partial_\mathbf{q}\mathbf{c}$ is the Jacobian matrix of constraints (partial derivative of the constraint vector $\mathbf{c}$ with respect to generalized coordinates $\mathbf{q}$), $\ddot{\mathbf{q}}$ is the vector of generalized accelerations (second derivative of $\mathbf{q}$ with respect to time $t$), $\boldsymbol{\lambda}$ is the vector of Lagrange multipliers, $\mathbf{f}^E$ is the vector of generalized applied forces, $\mathbf{f}^V$ is the vector of centrifugal forces, and $\mathbf{f}^D$ is the vector absorbing quadratic-velocity terms remaining after double time-differentiation of constraint equations.

This paper investigates the system composed of two bodies, therefore, the vector $\mathbf{q}$ of generalized coordinates includes two particular vectors $\mathbf{q}_1$, $\mathbf{q}_2$ of body coordinates. Equations (\ref{eq:mod_1}) are written in general notation, and their members acquire specific forms depending on the used multibody formalism. This work utilizes absolute coordinates together with Euler angles, and so the vector $\mathbf{q}_i$ of generalized coordinates describing the position and the orientation of each body ($i$ is the body index) is defined as $\mathbf{q}_i =
\begin{bmatrix}
    X_i & Y_i & Z_i & \psi_i & \vartheta_i & \varphi_i
\end{bmatrix}^\mathrm{T},$ where $X_i, Y_i, Z_i$ are absolute coordinates of the referential point $O_i$ and $\psi_i, \vartheta_i, \varphi_i$ is the sequence of Euler angles called the precession, the nutation, and the spin.


Expressions of $\mathbf{M}$, $\mathbf{f}^E$, $\mathbf{f}^V$ can be found in many publications dedicated to multibody dynamics, see e.g. \cite{Shabana2010}. However, the vector $\mathbf{c}$ of kinematic constraints, which derivatives form both $\partial_\mathbf{q}\mathbf{c}$ and $\mathbf{f}^D$, depends on the specific multibody system. Body 1 can perform only one primitive motion, which is the rotation with respect to the global axis $Z$, all other motions are not allowed. These kinematic restrictions can be mathematically formulated by five constraint equations $X_1 - X_1^0 = 0$,\;~$Y_1 - Y_1^0 = 0$,\; $Z_1 - Z_1^0 = 0$,\; $\psi_1 - \psi_1^0 = 0$,\; $\vartheta_1 - \vartheta_1^0 = 0$, where the superscript~$\cdot^0$ denotes the initial value of the particular coordinate.



Body 2 is constrained as well. Its motion is restricted by three cables that create the kinematic connection between the two bodies. With the assumption that all cables are always tightened and rigid, the distance between the ending points of the particular cable is constant, and corresponding constraint equations acquire the form of $||\mathbf{r}^A_1~-~\mathbf{r}^A_2||~-~l^A = 0$, $||\mathbf{r}^B_1~-~\mathbf{r}^B_2||~-~l^B~=~0$, $||\mathbf{r}^C_1~-~\mathbf{r}^C_2||~-~l^C~=~0$, where $l^A$, $l^B$, $l^C$ are lengths of cables, $\mathbf{r}^A_1$, $\mathbf{r}^B_1$, $\mathbf{r}^C_1$ are absolute position vectors of cables' end points on body 1, and $\mathbf{r}^A_2$, $\mathbf{r}^B_2$, $\mathbf{r}^C_2$ are position vectors of cables' ending points on body 2. The operator $||\cdot||$ represents the Euclidean norm of the vector.

The first set of constraints is already in a very simple form easy for subsequent processing since they are formulated only in terms of generalized coordinates. 
On the other hand, the second set of constraints includes kinematic variables, positions of specific points, that need to be expressed as functions of generalized coordinates. The position vector of an arbitrary point $M_i$ on body $i$ can be defined as
\begin{equation}
    \mathbf{r}^M_i = \mathbf{R}_i(\mathbf{q}_i) + \mathbf{T}_i(\mathbf{q}_i)\overline{\mathbf{u}}_i,
\label{eq:mod_5}
\end{equation}
where $\mathbf{R}_i = [X_i, Y_i, Z_i]^\mathrm{T}$ is the position vector of the referential point $O_i$, $\mathbf{T}_i$ denotes the transformation matrix from the local to the global coordinate system, which is dependent on the Euler angles $\psi_i, \vartheta_i, \varphi_i$, and $\overline{\mathbf{u}}_i$ is the position vector of the point $M_i$, which remains constant. Using general formula (\ref{eq:mod_5}), it is possible to express all six position vectors in constraint equations in terms of generalized coordinates.

When all terms in constraint equations are expressed, the constraint vector $\mathbf{c}(\mathbf{q})$ is created. Each element of the column vector corresponds to the left-hand side of the corresponding constraint equation. It is clear that differentiation of $\mathbf{c}$ is needed because there is both the Jacobian matrix $\partial_\mathbf{q}\mathbf{c}$ and the term $\mathbf{f}^D = \mathbf{f}^D(\partial_\mathbf{q}\mathbf{c},\partial^2_{t\mathbf{q}}\mathbf{c},\partial^2_{\mathbf{q}\mathbf{q}}\mathbf{c},\partial^2_{tt}\mathbf{c})$
in DAEs (\ref{eq:mod_1}). 

Correct symbolic derivation can be quite challenging; however, it is not necessary to perform the algebraic work manually, as many useful tools are now available to assist such analytical calculations (e.g., MATLAB Symbolic Math Toolbox). The model can also be assembled semi-automatically using state-of-the-art software tools for physical network acausal modeling, such as MATLAB Simscape, OpenModelica or Dymola, using the above explained assumptions.

As discussed later in the text, it is important to introduce two quantities that capture two basic motions performed by body 2 with respect to a motionless observer connected to the frame. The first dominant motion is horizontal torsion around the global axis $Z$, which is represented by the angle $\varepsilon_2$ oriented from the fixed plane $XZ$ to the moving plane $P_2Z$ defined by the global axis $X$ and the point $P_2$. This angle is depicted in Figure \ref{fig_mat_model_2D} {and can be expressed as 
\begin{equation}
    \varepsilon_2 = \frac{\pi}{2} - \cos^{-1} \left(  \frac{{(\mathbf{s}_2^P)}^\mathrm{T} \mathbf{n}^{XZ}}{||\mathbf{s}_2^P||}  \right),
\label{eq:mod_1b}
\end{equation}
where $\mathbf{s}_2^P = \begin{bmatrix}\mathbf{r}_2^P(1) & \mathbf{r}_2^P(2) & 0 \end{bmatrix}^\mathrm{T}$ is the vector pointing from the axis $Z$ to the point $P_2$ and $\mathbf{n}^{XZ} = \begin{bmatrix} 0 & 1 & 0\end{bmatrix}^\mathrm{T}$ is the unit normal vector of the plane $XZ$.
The second dominant motion is vertical nutation from the global axis $Z$. It is represented directly by the nutation angle $\vartheta_2$ visualized in Figure \ref{fig_mat_model_3D}. Furthermore, the actuated motor position is denoted as $\alpha$. Because the revolute joint between the motor and body~1 is ideal (rigid and with no clearance), it simply applies that the actual rotation of body {1} defined by the spin {$\varphi_1$} is identical to the actuated motor position $\alpha$, i.e., ${\varphi_1}(t) = \alpha(t)$.




\subsection{Local linearization}

The model (\ref{eq:mod_1}) can be transformed from the DAE form into ODEs (ordinary differential equations) by eliminating the Lagrange multipliers as follows
\begin{equation}
\begin{split}
\ddot{\mathbf{q}} = & \mathbf{M}^{-1} \cdot \lbrace \partial_\mathbf{q}\mathbf{c}^\mathrm{T} \cdot (\partial_\mathbf{q}\mathbf{c} \cdot\mathbf{M}^{-1} \cdot\partial_\mathbf{q} \mathbf{c}^\mathrm{T})^{-1} \\
  & \cdot  [\mathbf{f}^D - \partial_\mathbf{q}\mathbf{c} \cdot\mathbf{M}^{-1}\cdot (\mathbf{f}^E+\mathbf{f}^C)] + \mathbf{f}^E+\mathbf{f}^C \rbrace.
\end{split}
\label{eq:mod_2b}
\end{equation} 
Subsequently, this nonlinear model can be linearized around the stable equilibrium $\dot{\mathbf{{q}}}_i=0, \ddot{\mathbf{{q}}}_i=0, \dot{\alpha}=0$
with the suspended belt hanging steady without any movement.  Assuming $\varepsilon_2$ and $\vartheta_2$ to be the output variables, a linear time-invariant (LTI) model is obtained in the form of a transfer function matrix
\begin{align}
&\begin{bmatrix}
E_2 (s)\\
\Theta_2 (s)
\end{bmatrix} = \begin{bmatrix}
P_\varepsilon (s)\\
P_\vartheta (s)
\end{bmatrix}A(s)=[\mathbf{P_f}(s)+\mathbf{P_r}(s)]A(s),\nonumber \\
&\mathbf{P_f}(s)=\begin{bmatrix}
\frac{b_{1}s+b_{2}}{s^2+2\xi_1\omega_1s+\omega_1^2}+\frac{b_{3}s+b_{4}}{s^2+2\xi_2\omega_2s+\omega_2^2}\\
\frac{b_{5}s+b_{6}}{s^2+2\xi_1\omega_1s+\omega_1^2}+\frac{b_{7}s+b_{8}}{s^2+2\xi_2\omega_2s+\omega_2^2}
\end{bmatrix},
\end{align}
where $E_2 (s), \Theta_2 (s), A(s)$ denote the Laplace images of the belt torsion  and nutation angles $\varepsilon_2,\vartheta_2$ and actuated motor position $\alpha$, respectively, and $P_\varepsilon,P_\vartheta$ denote the dynamic contribution of the input to the two outputs, $\mathbf{P_f}$ expresses the contribution of the oscillatory modes and $\mathbf{P_r}$ contains the rigid body modes and other residual dynamics.


The numerator coefficients $b_i$ depend on the physical parameters of the system. However, only the natural frequencies and damping factors $\omega_{1,2},\xi_{1,2}$ matter for the proposed design of the input shaping filter, described in Section V.


The term $\mathbf{P_f}$ describes the input-output contribution of the two dominant modes of oscillatory motion the system typically exhibits (Fig. \ref{system_hw_real}). The \emph{torsional} bending mode, mainly captured by the introduced output angle $\varepsilon_2$, is caused by a natural tendency of the belt to maintain its steady-state angular velocity while the actuator is accelerating, causing an oscillatory motion around the belt's axis of symmetry. This mode is always present for most physically meaningful combinations of parameters, unless there is an excessive amount of damping added artificially in the model. The second \emph{nutation} mode exhibits itself as a swinging pendulum-like translational motion, shifting the belt in a mostly horizontal manner. This behavior arises due to the centrifugal forces resulting from the actuator-induced motion, in case there is a shift of the hanging load's center of gravity with respect to the axis of symmetry. In practice, this asymmetry is always present due to the heavy metal buckle. The nutation angle is directly captured in the second output $\vartheta_2$ of the system model.

\section{Vibration damping via input-shaping method}


For the effective attenuation of both bending modes in the system dynamics, an input-shaping method is proposed, due to a~few inherent advantages it offers for the Flying belt application:

\begin{itemize}
\item The trajectory reference filter operates in a feedforward configuration between the vision tracking system and the actuator, eliminating the need for direct oscillation measurement and simplifying the system. Although the belt can in principle be tracked with the same camera observing the visitors passing by, the issues of limited update rate, computational complexity and closed-loop stability would arise.
\item Only the knowledge of the system's resonant frequencies is required, rather than a full dynamic model. The resonances' location is examined experimentally by measuring oscillation periods for the two normal modes of vibration. 
\item Robustness to modeling uncertainties in bending mode frequencies is incorporated into the filter design.
\item The filter design is cast as a convex optimization problem with a unique global solution, efficiently solvable by standard numerical methods.
\end{itemize}

\subsection{Zero vibration input shaping fundamentals}
We consider a class of so-called \emph{lumped delay} shaping filters with a finite impulse response $h_s$ taking the form of a weighted Dirac pulse series
\begin{equation}
h_s(t)= \suma{i=1}{n}A_i\delta(t-t_i);~t_i<t_{i+1}, y(t)=\suma{i=1}{n}A_i u(t-t_i)\label{eq:shaper_cont}
\end{equation}
where $\delta$ denotes the Dirac delta distribution (unit impulse), $A_i$ stands for the amplitudes or weighting factors, $t_i$ define the time shift of the pulses and $u(t),y(t)$ denote shaper input and output signals.

A common way of designing input shapers is via the \emph{shaper sensitivity function} $V(\omega,\xi)$ \cite{singer1990preshaping}

\begin{align}
&V(\omega,\xi)=e^{-\xi\omega t_n} \sqrt{C(\omega,\xi)^2+S(\omega,\xi)^2};\\
 &C(\omega,\xi)\overset{\Delta}{=}\sum \limits_{i=1}^{n} A_i e^{\xi \omega t_i} \cos(\omega_{d} t_i),\nonumber\\
 &S(\omega,\xi)\overset{\Delta}{=}\sum \limits_{i=1}^{n} A_i e^{\xi \omega t_i} \sin(\omega_{d} t_i),~\omega_d=\omega\sqrt{(1-\xi^2)},\nonumber
\label{eq:zvn_ar}
\end{align}
expressing a relative amount of excited residual oscillations of a particular oscillatory mode parameterized by its natural frequency $\omega$ and damping factor $\xi$.
Minimum, i.e. zero, level of oscillation for a particular mode can be achieved by setting
\begin{equation}\label{eq:zvn_cond}
V(\omega_k,\xi_k)=0,\frac{\partial V(\omega_k)}{\partial \omega_k}{=}0,~\frac{\partial V(\omega_k)}{\partial \xi_k}{=}0,
\end{equation}
for a k-th oscillatory mode. The first equality in (\ref{eq:zvn_cond}) is often designated as \emph{zero vibration condition} since it ensures a~complete elimination of residual vibrations in a finite time $t_n$ that corresponds to the finite length of the shaper impulse function. The derivatives in $(\ref{eq:zvn_cond})$ address the issue of uncertainty in $\omega_k,\xi_k$ by enforcing a flat shape of the sensitivity function in the vicinity of assumed modal parameters, making the shaper more robust to modeling errors.




\subsection{Dual-mode robust input shaper design}

In our application to the Flying belt motion control problem, we combine and elaborate previous theoretical results published in \cite{cole2012class, goubej2020frequency} to derive a compact shaper design algorithm allowing for automated synthesis of the input shaping filter precompensating both resonant modes of the system while addressing both robustness to uncertainty and smoothness of the generated motion trajectories. 

The starting point is the estimated location of the two dominant resonance modes expressed by their natural frequencies and damping factors, defining a parametric model set $\mathcal{M}\overset{\Delta}{=}\{\omega_1,\xi_1,\omega_2,\xi_2\}$ with the parameters either extracted from the first-principles model introduced in Section IV, or from a simple identification experiment with the physical belt.

Since we target digital implementation, a discrete-time version of the shaping filter (\ref{eq:shaper_cont}) is assumed. By setting $0=t_{1}<t_{2}<...<t_{n}$ and $t_{i+1}-t_{i}=T_s, i=2..n$ with $T_s$ being a chosen sampling period, the shaper adopts a structure of a digital finite-impulse-response (FIR) filter which can be represented by a discrete transfer function in the Z-domain as
\begin{align} \label{eq:ISTFD}
{H}\left(z\right)&\overset{\Delta}{=}\frac{y(z)}{u(z)}=\underset{i=1}{\overset{n}{\sum}}A_{i}z^{-t_i/T_s},
{\bf h}\overset{\Delta}{=}[A_1, A_2, ..., A_n]^T,
\end{align}
where the integer $n$ defines the filter length, leading to a~corresponding set of shaper amplitudes which may be stored in the form of a vector $\bf h$ also defining the time-sequence of the filter's finite impulse response function $h(k\cdot T_s)=A(k+1);k=0..n-1$.

The shaper design procedure can be formulated as a constrained optimization problem in the form of 
\begin{equation}
 \min_{\mathbf{h}}\ f(\mathbf{h})~\mathrm{subject\ to }
\left\{
\begin{array}{@{}l@{}}
 \mathbf{A}_{\mathbf{eq}}(\mathcal{M})\mathbf{h}=\mathbf{b}_{\mathbf{eq}}(\mathcal{M}) ,\\
 \mathbf{-I}\mathbf{h}\leq\mathbf{0},\end{array}\right.\label{eq:optim}
\end{equation}
where $f(\mathbf{h})$ is an objective function, $\mathbf{A}_{\mathbf{eq}},\mathbf{b}_{\mathbf{eq}}$ are real matrices of compatible dimensions constructed from the zero vibration conditions (\ref{eq:zvn_cond}) and the second inequality constraint with the identity matrix $\mathbf{I}$ is introduced to enforce smooth filters with a monotonous step response.

\subsection*{Time-optimal solution via feasibility problem}
By omitting the objective function $f(\mathbf{h})$ in  (\ref{eq:optim}), the feasibility problem can be solved for a preselected filter length $n$ using state-of-the-art numerical methods for the linear programming (LP) problem. Generally speaking, lower values of $n$ are preferred in practice to minimize the shaper action time and the corresponding transient delay it introduces in the shaped signal.  A minimum feasible shaper duration is introduced as $n_{min}\overset{\Delta}{=}\min_{n \in \mathbb{Z}^+} \; n \quad \text{subject to} \quad \exists\, \mathbf{h} \in \mathbb{R}^n : \; \text{constraints}(\mathbf{h})$, which can be found e.g. by a simple bisection algorithm, as outlined in \cite{linprog}.


\subsection*{Quadratic optimal shaper-smoother}

Once the lower bound for the shaper duration $n_{min}$ was found, the objective function $f(\mathbf{h})$ in (\ref{eq:optim}) is chosen as $f(\mathbf{h})=\mathbf{h^T h}=\underset{\forall i}{\sum}{A_i}^2=\Vert H(z)\Vert _2^2$,
where $||.||_2$ denotes the H2 norm of a linear time-invariant system. This particular choice leads to several practical advantages, as discussed in \cite{goubej2020frequency}:
\begin{itemize}
    \item Minimization of shaper's H2 norm also  minimizes the average gain of shaper's amplitude frequency response function $||H(z)||_2^2=\frac{1}{2\pi}\overset{\pi/Ts}{\underset{-\pi/T_s}{\int}}|H(e^{j\omega T_s})|^2\mathrm{d}\omega$, thereby producing smoother motion trajectories that are easier to track by a physical actuator. A chance of exciting unmodelled dynamics, e.g. omitted higher bending modes, is also lower.


\item The quadratic cost function together with the linear constraints in (\ref{eq:optim}) constitute a convex \emph{quadratic programming problem} (QP) for which a global optimal solution can be found using SoA numerical methods, e.g. the interior point or active-set algorithms. The optimization is time-effective, allowing for the synthesis of filters with dimensions of $n\sim 10^3$ in the order of seconds.
\end{itemize}

The search for a suitable shaping filter is then constrained to a~finite set of the admissible duration parameter values $\{ n \in \mathbb{Z}^+ \mid n_{min} \leq n \leq  2\cdot n_{min}\}.$
The time-optimal solution obtained for $n=n_{min}$ typically leads to sparse filters with high-values of amplitudes $A_i$ concentrated in clusters separated by wide fields of zero entries. This allows for fast settling times, but may produce physically infeasible motion commands. On the other hand, the upper limit $2\cdot n_{min}$ prevents from the use of very long filters with sluggish response. 

A dimensionless scaling called \emph{smoothing factor} $\{\, s_f \in \mathbb{R} \mid 0 \leq s_f \leq 1 \,\}$ is introduced together with the mapping $n = n_{min} + \texttt{round}(s_f\cdot n_{min}),\label{eq:sf}$ to simplify shaper fine-tuning by the user, allowing him to find a sweet spot between the introduced delay and provided level of signal smoothing. 

The filter synthesis can be fully automated and is performed only once during system commissioning, while the real-time execution involves efficient discrete-time convolution, calculating the shaping filter output as a weighted sum of the current and past inputs (\ref{eq:shaper_cont}).

\begin{algorithm}[h]\label{alg}
\caption{Dual-mode robust shaper-smoother design}
\begin{algorithmic}[1]
    \REQUIRE Modal parameters $\mathcal{M}$, smoothing factor $s_f$ 
    \ENSURE Impulse function of the digital FIR filter $\mathbf{h}$ (\ref{eq:ISTFD})
    \STATE Construct equality and inequality constraints from the zero-vibration conditions for both oscillatory modes from (\ref{eq:zvn_cond},\ref{eq:optim})
    \STATE Find the minimal shaper duration $n_{min}$ by means of the feasibility problem (\ref{eq:optim}) using HiGHS LP solver
    \STATE Compute the required shaper duration $n$ from based on the user-specified smoothing factor $s_f$ to introduce the smoothing functionality 
    \STATE Find the quadratic optimal shaper $\mathbf{h}(n,\mathcal{M})$ by solving the QP problem in (\ref{eq:optim}) for $f(\mathbf{h})=\mathbf{h^T h}$ (DAQP solver)
\end{algorithmic}
\end{algorithm}

\section{Numerical simulation}
The proposed motion control scheme was validated by means of a numerical simulation using the first principle model derived in Section IV. Table \ref{tab:parametry} shows the set of physical parameters chosen to match an existing physical setup. Figure \ref{fig_frf} displays the frequency response function of the linearized dynamic model, with clear resonance peaks caused by the two dominant oscillatory modes.

{
\setlength{\abovecaptionskip}{-0pt}
\begin{figure}[!t]
\centering
\includegraphics[width=0.485\textwidth]{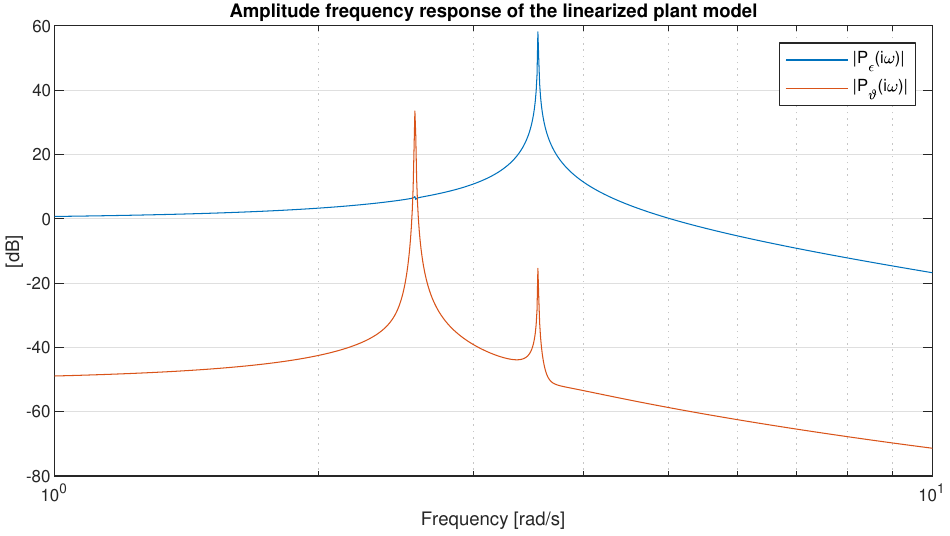}
\caption{Amplitude frequency response function of the linearized model}
\label{fig_frf}
\vspace{-0.5cm}
\end{figure}
}

A point-to-point positioning scenario was simulated, aiming at changing the orientation of the suspended belt by $180$ degrees as fast as possible. This emulates the longest maneuver that may occur in practice due to the cyclical configuration of the rotary motion stage. 

{
\setlength{\abovecaptionskip}{-2pt}
\begin{figure*}[t]
\centering
\includegraphics[width=0.85\textwidth]{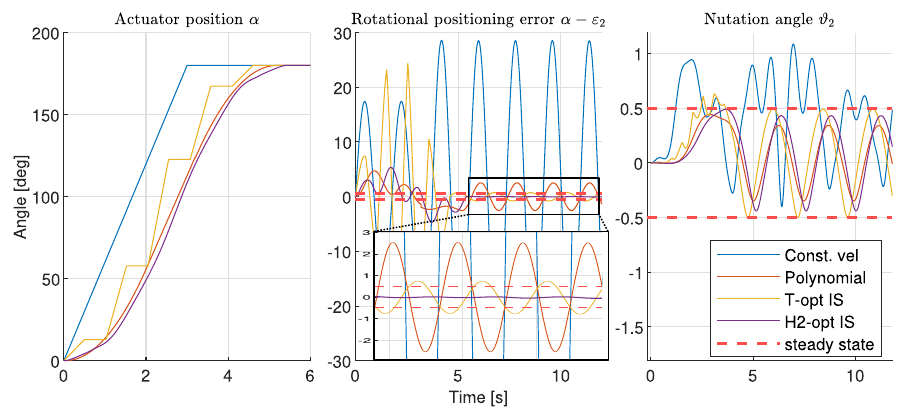}
\caption{Point-to-point positioning scenario using different motion planning algorithms - constant velocity profile (blue), jerk-limited 3rd degree polynomial profile (red), time-optimal input shaper (yellow), $H_2$-optimal shaper-smoother (purple)}
\label{fig:fig_simul}
\end{figure*}
}

\begin{table}[h]
    \centering
    \begin{tabular}{|l|c|}
        \hline
        Physical parameter    & Value        \\
        \hline
        Cable lengths $l^{A,B,C}$ [m]   & 1.5           \\
        Motion stage radius $r_1$ [m]          & 0.32            \\
        Belt radius $r_2$ [m]           & 0.15         \\
        Belt weight $m_2$ [kg]       & 0.147             \\
        Belt CoM distance $c_2$[m] & 0.015            \\
        Belt inertia matrix $\mathbf{M_2}$[kg.m$^2$] & diag(0.0015,0.0018,0.0033)            \\
        Resonance frequencies $\omega_{1,2}$ [rad/s] & 2.58,~3.55            \\
        \hline
    \end{tabular}
    \vspace{2pt}
    \caption{Model parameters}
    \label{tab:parametry}
\vskip -0.4cm
\end{table}

Four different motion planning strategies are compared in Fig. \ref{fig:fig_simul}. The first one uses a simple linear position rate change limiter, achieving a constant velocity motion profile and a~significant amount of belt oscillations. The second strategy involves computing an S-shaped 3rd-degree piecewise polynomial trajectory with constrained velocity, acceleration and jerk; a shape commonly used in industrial motion systems. The oscillations are mitigated to some extent at the cost of prolonging the actuator transient time. Notable level of torsional oscillations still remains. The next one is a position profile planned by means of the time-optimal input shaper, which is able to reduce both the level of belt residual oscillations and actuator traversal time when compared to the polynomial trajectory. A~potential drawback is the step-wise change in the commanded position, which may excite unmodelled dynamics or cause issues with reference tracking using a~physical actuator. The use of the $H_2$-optimal shaper designed for $s_f=0.15$ mitigates these issues due to the introduced smoothing functionality, showing a significantly lower level of torsional oscillations compared to the conventional polynomial smoother, while achieving similar settling time.

The third plot in Fig. \ref{fig:fig_simul} reveals that a small residual deviation in the nutation angle still remains, even with the input shaping method employed. This is an inherent result of the nonlinear system dynamics. It is impossible to rotate the system without introducing centrifugal forces acting on the asymmetrically suspended load with the center of gravity shifted with respect to the axis of rotation. The level of nutation can be further reduced by increasing the shaper action time via the introduced smoothing factor $s_f$, allowing for finding a suitable trade-off between system responsiveness and the level of excited oscillations; see the appended movie using the link below, comparing the different motion planning strategies using an identical actuator maneuver time. 

\begin{table}[h]
    \centering
    \begin{tabular}{|l|c|c|c|c|}
        \hline
        Perf. measure & Const. vel & Polynomial & T-opt IS & $H_2$-opt\\
        \hline
        Err. torsion [pk-pk] & 57.2      & 5.07      & 1.56      & 0.0941      \\
        Err. torsion [rms]        & 20.1       & 1.77       & 0.54       & 0.0484       \\
        Err. nutation [pk-pk]          & 1.52    & 0.688    & 0.988     & 0.869    \\
        Err. nutation [rms]     & 0.57       & 0.241        & 0.336        & 0.304        \\
        Trans. time act [s]         & 3       & 5.3       & 4.6       & 5.4       \\
        Trans. time belt [s]         & $>60$       & $>60$       & $\approx 4.6$       & 5.4       \\
        
        \hline
    \end{tabular}
    \vspace{2pt}
    \caption{Performance in the PTP positioning scenario}
    \label{tab:stats}
\end{table}

Table \ref{tab:stats} summarizes the statistics of the achieved belt positioning errors by means of the peak-peak and root-mean-square errors of the residual torsion $\alpha-\varepsilon_2$ and nutation angles. Although the conventional command smoothing reduces the oscillations to some extent, a notable residual belt motion remains, as indicated by the transient time required to settle both the measured angles below the level of approximately 0.5~degrees, denoting steady state. Moreover, the performance achieved with polynomial motion commands varies significantly with the desired length of the rest-to-rest motion and the tuning of the maximum acceleration and jerk parameters is highly non-intuitive. This is because the timing of the acceleration phases in the actuator movement must match the actual phase of the belt oscillation correctly to attenuate its amplitude. On the other hand, the proposed input shaping method improves the motion tracking performance consistently for any desired trajectories, due to the principle of convolution embedded in its design and operation.

\section{Experimental results}


This work was presented in September of 2024 during Rafael Lozano-Hemmer's solo show, \textit{Caressing the Circle}, at bitforms gallery in New York. This is the second time the piece has been presented at the gallery in the past 20 years. bitforms is one of the longest running media art galleries in the United States and the show was well received by the art community in New York and abroad. Approximately 350 visitors came to the opening reception to view the work. 

Functionality of the interactive artwork is demonstrated in a~movie provided at the following link: \href{https://bit.ly/flyingbelt25}{https://bit.ly/flyingbelt25}

\section{Conclusion}

The refurbishment of the control system resulted in enhanced modularity, greater cost-effectiveness, and simplified scalability of the overall solution. The upgraded motion control algorithms significantly reduced residual oscillations, thereby improving both the responsiveness and interactivity of the kinetic installation. Future research may focus on exploring various feedback control architectures, either by combining the camera with a belt motion estimator or by employing load cells to measure tension forces in the suspended cables. The main goal is to further improve motion performance and robustness, especially in the presence of external disturbances.

\section*{Acknowledgments}
This project was co-funded by the European Union under the project Robotics and Advanced Industrial Production (reg. no. CZ.02.01.01/00/22\_008/0004590). The support is gratefully acknowledged.



 
%

\bibliographystyle{IEEEtran}

\bibliography{references}

@software{accelstepper,
  author = {Patrick Wasp},
  title = {{AccelStepper}},
  url = {https://github.com/waspinator/AccelStepper},
  version = {1.64},
  date = {2022},
}

@online{standards,
  title = {{Rafael Lozano-Hemmer - Standards and Double Standards}},
  url = {https://www.lozano-hemmer.com/standards_and_double_standards.php},
  urldate = {2025-06-27}
}

@online{across,
  title = {{Automation, Control and Robotic Systems Group (ACROSS), NTIS Research Center, Pilsen, Czechia}},
  url = {https://www.ntis.zcu.cz/en/Research/Teams/ACROSS/index.html},
  urldate = {2025-06-27}
}

@book{Schoffer63,
  title={Nicolas Sch\"{o}ffer},
  author={Sch\"{o}ffer, N.},
  year={1963},
  publisher={Neuchatel: Editions du Griffon}
}

@article{goubej2020frequency,
  title={{Frequency weighted H2 optimization of multi-mode input shaper}},
  author={Goubej, Martin and Vyhl{\'\i}dal, Tom{\'a}{\v{s}} and Schlegel, Milo{\v{s}}},
  journal={Automatica},
  volume={121},
  pages={109202},
  year={2020},
  publisher={Elsevier}
}

@INPROCEEDINGS{linprog,

  author={Van den Broeck, L. and Pipeleers, G. and De Caigny, J. and Demeulenaere, B. and Swevers, J. and De Schutter, J.},

  booktitle={2008 10th IEEE International Workshop on Advanced Motion Control}, 

  title={A linear programming approach to design robust input shaping}, 

  year={2008},

  volume={},

  number={},

  pages={80-85},

  doi={10.1109/AMC.2008.4516045}}

@Article{cole2012class,
  Title                    = {A class of low-pass {FIR} input shaping filters achieving exact residual vibration cancelation},
  Author                   = {Cole, Matthew OT},
  Journal                  = {Automatica},
  Year                     = {2012},
  Number                   = {9},
  Pages                    = {2377--2380},
  Volume                   = {48},

  Publisher                = {Elsevier}
}

@Article{singer1990preshaping,
  Title                    = {Preshaping command inputs to reduce system vibration},
  Author                   = {Singer, Neil C and Seering, Warren P},
  Journal                  = {Journal of dynamic systems, measurement, and control},
  Year                     = {1990},
  Number                   = {1},
  Pages                    = {76--82},
  Volume                   = {112},

  Publisher                = {American Society of Mechanical Engineers}
}

@article{cuan2021,
  title={Output: choreographed and reconfigured human and industrial robot bodies across artistic modalities},
  author={Cuan, Catie},
  journal={Frontiers in Robotics and AI},
  volume={7},
  year={2021},
  publisher={Frontiers Media SA}
}

@article{kac1997digital,
  title={Digital reflections: the dialogue of art and technology},
  author={Kac, E},
  journal={Art Journal},
  volume={56},
  number={3},
  pages={60--67},
  year={1997}
}

@inproceedings{Qin_2025, series={CHI ’25},
   title={Encountering Robotic Art: The Social, Material, and Temporal Processes of Creation with Machines},
   booktitle={Proceedings of the 2025 CHI Conference on Human Factors in Computing Systems},
   publisher={ACM},
   author={Qin, Yigang and Li, Yanheng and Cheon, EunJeong},
   year={2025},
   month=apr, pages={1–18},
   collection={CHI ’25} }

@article{st2019robotic,
  title={Robotic art comes to the engineering community [Art and Robotics]},
  author={St-Onge, David},
  journal={IEEE Robotics \& Automation Magazine},
  volume={26},
  number={3},
  pages={103--104},
  year={2019},
  publisher={IEEE}
}

@ARTICLE{Zhang14,

  author={Zhang, Tan and Backstrom, Kirk and Prince, Richard and Liu, Changli and Qian, Zhiqin and Zhang, Dan and Zhang, Wenjun},

  journal={IEEE Robotics \& Automation Magazine}, 

  title={Robotic Dynamic Sculpture: Architecture, Modeling, and Implementation of Dynamic Sculpture}, 

  year={2014},

  volume={21},

  number={3},

  pages={96-104},

  doi={10.1109/MRA.2014.2312842}}

@article{gomez2021robot,
  title={The robot is present: Creative approaches for artistic expression with robots},
  author={Gomez Cubero, Carlos and Pekarik, Maros and Rizzo, Valeria and Jochum, Elizabeth},
  journal={Frontiers in Robotics and AI},
  volume={8},
  pages={662249},
  year={2021},
  publisher={Frontiers Media SA}
}

@article{beltramello2020artistic,
  title={Artistic robotic painting using the palette knife technique},
  author={Beltramello, Andrea and Scalera, Lorenzo and Seriani, Stefano and Gallina, Paolo},
  journal={Robotics},
  volume={9},
  number={1},
  pages={15},
  year={2020},
  publisher={MDPI}
}

@article{robinson2023robotic,
  title={Robotic vision for human-robot interaction and collaboration: A survey and systematic review},
  author={Robinson, Nicole and Tidd, Brendan and Campbell, Dylan and Kuli{\'c}, Dana and Corke, Peter},
  journal={ACM Transactions on Human-Robot Interaction},
  volume={12},
  number={1},
  pages={1--66},
  year={2023},
  publisher={ACM New York, NY}
}

@book{Shabana2010,
  title={Computational dynamics},
  author={Shabana, A. A.},
  year={2010},
  publisher={Wiley}
}

@book{riskin,
author = {Riskin, Jessica},
title = {Genesis Redux: Essays in the History and Philosophy of Artificial Life},
year = {2007},
isbn = {0226720802},
publisher = {University Of Chicago Press}
}


 




\vfill

\end{document}